\documentclass[sigconf]{acmart}

\AtBeginDocument{%
	}

\setcopyright{acmlicensed}
\copyrightyear{2018}
\acmYear{2018}
\acmDOI{XXXXXXX.XXXXXXX}
\acmConference[Conference acronym 'XX]{Make sure to enter the correct
	conference title from your rights confirmation email}{June 03--05,
	2018}{Woodstock, NY}
\acmISBN{978-1-4503-XXXX-X/2018/06}

\usepackage{multirow}
\usepackage{tabularx}
\usepackage{color}
\usepackage{xcolor}
\usepackage{booktabs}

\usepackage[vlined,boxed,commentsnumbered,linesnumbered,ruled]{algorithm2e}
\usepackage{amsmath}
\usepackage{tabularx}
\usepackage{adjustbox}
\usepackage{multirow}
\usepackage{colortbl}
\usepackage{listings}
\usepackage{pifont}

\definecolor{green}{RGB}{113,165,55}
\definecolor{blue}{RGB}{1,158,213}
\definecolor{red}{RGB}{220,10,10}
\definecolor{LightGreen}{HTML}{d4edda}  
\definecolor{LightBlue}{HTML}{d1ecf1}   
\usepackage[capitalize]{cleveref}
\Crefname{section}{Section}{Sections}
\Crefname{table}{Table}{Tables}
\Crefname{figure}{Figure}{Figures}
\Crefname{equation}{Equation}{Equations}
\Crefname{paragraph}{Paragraph}{Paragraphs}

\setcopyright{none}
\renewcommand\footnotetextcopyrightpermission[1]{}
\begin{document}
	
	\title{Selective Positional Allocation for Noise-Robust Query Retrieval in Small-Object Detection}
	
	\author{Yangchen Zeng}
	\affiliation{%
		\institution{Southeast University}
		\city{Nanjing}
		\country{China}
	}

	\begin{abstract}
			In aerial and remote-sensing imagery, detecting tiny targets is frequently hampered less by the representational capacity of the backbone network than by the manner in which positional priors are distributed across ill-defined foreground and background regions. We introduce \textbf{SPAR} (Selective Positional Allocation and Retrieval), a detection framework that re-conceptualizes positional cues as an allocatable resource: they are intensified at semantically meaningful locations while responses driven by scene clutter are actively attenuated. At the heart of SPAR lies \textbf{Selective Positional Embedding (SPE)}, an interpretable embedding mechanism paired with an activation-bar visualization that facilitates direct inspection and hyper-parameter tuning. The encoder branch, termed \textbf{MOHFE}, fuses heatmap-modulated positional signals with category-specific and box-level evidence to yield more compact multi-scale features; the decoder branch, termed \textbf{HIDQ}, employs a foreground gating operation to discard background-dominated candidates prior to query decoding. To cope with the extreme sparsity of small-object evidence, we additionally design \textbf{DPSConv} (Dual-Path Sampling Convolution), which augments retrieval-oriented representations through two geometrically complementary sampling routes. Notably, all heatmap supervision is derived from existing detection objectives and consumed exclusively during training, requiring no auxiliary heatmap annotations and leaving inference as an ordinary forward pass with no additional gradient overhead. Comprehensive evaluations on DOTA, DIOR, VisDrone, PASCAL VOC, NWPU VHR-10, and AITOD demonstrate the effectiveness of the proposed design: relative to the baseline, SPAR attains an improvement of up to 9.1 percentage points, shortens the decoder from eight layers to three, and reduces model parameters by 59.4\% (66.3M versus 163M).

		\end{abstract}

	
	\ccsdesc[500]{Computing methodologies~Computer vision problems}

	\keywords{Selective positional allocation, Query retrieval, Small object detection, Efficient decoder design}
	
	\maketitle

	\section{Introduction}
	\label{sec:intro}

	Detecting objects that occupy only a handful of pixels remains one of the most stubborn challenges in computer vision, and the difficulty is amplified in aerial and remote-sensing imagery where tiny targets are readily submerged in complex background textures. Such imagery is typified by wide scale disparities, highly congested instance layouts, and a drastic imbalance between foreground and background pixels, each of which independently complicates localization and classification~\cite{wei2024review, yang2022scrdet++, li2025efficient, hou2020mdpcaps}. Meanwhile, real-world deployment scenarios increasingly demand detectors whose accuracy does not collapse when imaging conditions or scene densities vary~\cite{cheng2023towards, cui2025efficient, akyon2022sahi}.

	Query-based detectors built upon the Transformer architecture have gained considerable traction in the small-object regime, owing to their global receptive fields and set-prediction paradigm that obviates dense anchor design and heuristic post-processing~\cite{zhao2024detrs, zhang2022dino,zeng2025hmpe}. Nevertheless, their effectiveness in clutter-heavy aerial scenes is fundamentally gated by the caliber of queries fed into the decoder. A considerable number of existing query-construction pipelines fail to fully exploit readily available detection signals---spatial coordinates, category confidences, and box regression gradients---particularly when annotated data are scarce. The consequence is a misalignment between positional encodings and detection semantics, whereby regions carrying useful information are not sufficiently emphasized for precise small-object localization~\cite{meng2021conditional, li2022dn}.
	
	A complementary shortcoming arises from the overwhelming background-to-foreground ratio: queries are readily polluted by activations originating from background regions. Areas of low semantic quality can nevertheless produce non-negligible embeddings, yet few detectors explicitly intervene to suppress this background-borne noise prior to decoding, thereby degrading both localization fidelity and classification accuracy~\cite{zhang2022dino, li2022dn, zhang2026center}. This phenomenon also engenders an efficiency bottleneck, since the decoder is compelled to allocate multiple layers to rectifying suboptimal queries, inflating both training and inference costs as self-attention complexity scales with token count and stack depth~\cite{zhao2024detrs}. The predicament is especially acute for small objects, whose discriminative features are inherently sparse and fine-grained. In the absence of a mechanism dedicated to amplifying such sparse discriminative patterns, query-relevant representations stay brittle under clutter and fail to generalize across domains~\cite{wei2024review, di2025toward}.

	Motivated by these observations, we put forward \textbf{SPAR} (Selective Positional Allocation and Retrieval), a framework that investigates a foundational question in query-based detection: where should positional cues be placed? Instead of uniformly broadcasting positional encodings across every spatial location, SPAR concentrates them in foreground-salient regions while actively dampening them in background-dominated zones. This selective distribution tightly couples positional priors with semantic and geometric evidence, elevating the quality of decoder inputs and endowing the detector with greater resilience to scene clutter. A cornerstone of SPAR is \textbf{Selective Positional Embedding (SPE)}, complemented by an activation-bar visualization that renders the embedding behavior directly inspectable and tunable.

	Leveraging SPE as a shared allocation signal, we devise two synergistic modules: \textbf{MOHFE} (Multi-Scale ObjectBox-Heatmap Fusion Encoder) and \textbf{HIDQ} (Heatmap-Induced Discriminative Query retrieval). MOHFE produces more refined multi-scale encoder features by interweaving category and geometric cues with selectively gated positional embeddings, thereby sharpening the correspondence between spatial priors and detection semantics. HIDQ subsequently assembles a compact set of discriminative queries and employs a foreground gate to eliminate background-dominated candidates, which diminishes the necessity for iterative query refinement and permits a substantially shallower decoder. The net effect is a simultaneous gain in efficiency and robustness.

	To confront the extreme feature sparsity exhibited by intricate small targets, we additionally devise \textbf{DPSConv} (Dual-Path Sampling Convolution), which captures sparse yet informative local patterns and feeds them into the retrieval pipeline, thereby enriching the representational basis upon which high-quality queries are constructed.

	The principal contributions of this work are enumerated below:
	\begin{itemize}
		\item We introduce \textbf{SPE}, an interpretable selective positional embedding mechanism accompanied by an activation-bar visualization for transparent analysis and convenient tuning of the allocation behavior.
		\item We design \textbf{MOHFE} and \textbf{HIDQ} to attain cleaner encoder representations and more discriminative query retrieval, collectively enabling a reduction of decoder depth from eight layers to three.
		\item We incorporate \textbf{DPSConv} to reinforce sparse feature extraction, achieving a 59.4\% parameter reduction (66.3M versus 163M) while accelerating both training and inference.
		\item We conduct thorough empirical validation on five primary benchmarks plus an additional AITOD ablation configuration, demonstrating consistent performance gains and favorable scalability.
		
	\end{itemize}

	\section{Related Work}
	
	Small-object detection can be viewed from two complementary angles: strengthening the visual features that describe tiny targets, and improving the decision process that turns those features into predictions. In aerial scenes, both are difficult at the same time because objects occupy few pixels, scene context is cluttered, and background responses often dominate the signal. Existing work therefore tends to focus either on more discriminative backbone features or on more reliable query/decoder design.
	
	\subsection{CNN-based Small Object Detection}
	CNN detectors remain a standard choice because pyramid-shaped representations naturally encode spatial detail at multiple scales. One-stage methods such as SSD~\cite{liu2016ssd}, RetinaNet~\cite{lin2017focal}, and the YOLO family~\cite{redmon2018yolov3,zyc2022behavior} emphasize efficiency, whereas two-stage systems such as Faster R-CNN~\cite{ren2015faster} and Mask R-CNN~\cite{he2017mask} refine proposals for stronger localization. For aerial imagery, many variants add extra machinery to recover tiny targets, including super-resolution enhancement~\cite{zhang2023superyolo} and comparative feature learning~\cite{cao2023pcnet}.
	
	Despite these improvements, CNN pipelines still rely heavily on dense candidate generation and NMS-style selection~\cite{hosang2017learning}. That design is vulnerable to scene-dependent thresholds and introduces redundancy when a high-resolution image contains many background patches but only a few useful targets~\cite{cai2018cascade, cui2025detection, ren2015faster, hou2025fedc}. In practice, the bottleneck is not only feature extraction, but also how to avoid spending computation on regions that are unlikely to matter.
	
	\subsection{Vision Transformer in Object Detection}
	Transformer detectors recast detection as set prediction, so the model learns a fixed number of queries instead of scanning dense candidate grids~\cite{carion2020end}. DETR and its successors, including Deformable DETR~\cite{zhu2020deformable}, Conditional DETR~\cite{meng2021conditional}, DINO~\cite{zhang2022dino}, and RT-DETR~\cite{zhao2024detrs}, combine convolutional feature extraction with transformer encoding and query-driven decoding. The main advantage is cleaner post-processing; the main weakness is that attention and query refinement become expensive when the input is large or the target is small.
	
	Recent work has therefore focused on reducing that cost or making the queries more informative. DynamicViT~\cite{rao2021dynamicvit} prunes tokens to save computation, CAM-based supervision has been used to assist localization under imbalance~\cite{ma2024coarse}, and sparse-attention designs target wide-field imagery by focusing on target-relevant regions~\cite{li2024sparseformer}. Related efficiency strategies include window attention~\cite{liu2021swin}, hierarchical or downsampling architectures~\cite{wang2021pyramid,wu2021cvt,zeng2026deepinterestgr,dong6413179hmpe}, and low-rank approximations~\cite{xiong2021nystromformer}.
	
	Even with these advances, query-based detectors remain fragile in cluttered small-object scenes: useful cues are sparse, background responses are strong, and the model may still devote too much decoding effort to poor queries. That observation motivates SPAR, which treats positional cues as something to allocate selectively rather than inject uniformly.
	
	\subsection{Positional Encoding and Heatmap-guided Query Construction}
	Positional encoding constitutes a foundational component in transformer-based detectors, as it supplies the spatial awareness that self-attention alone cannot capture. Early designs relied on fixed sinusoidal encodings~\cite{carion2020end}, while subsequent work explored learnable positional embeddings and conditionally generated spatial priors~\cite{meng2021conditional}. A particularly relevant line of research investigates heatmap-guided positional embedding, where detection-derived saliency maps are leveraged to modulate how positional cues are distributed across spatial locations. HMPE~\cite{zeng2025hmpe} demonstrates that heatmap-conditioned embeddings can substantially improve the efficiency of transformer-based small-object detection by aligning positional priors with object-centric regions. Building upon this insight, recent work further introduces noise-aware positional embedding for query retrieval~\cite{zeng2026learning}, which explicitly accounts for background interference when allocating positional resources. These studies collectively establish that the manner in which positional information is assigned---rather than merely whether it is present---plays a decisive role in query-based detection, a principle that SPAR extends through a fully differentiable selective allocation mechanism.
	
	\subsection{Multi-scale Representation and Cross-modal Feature Learning}
	Robust small-object detection hinges on the quality of multi-scale feature representations, since tiny targets must be resolved from high-resolution feature maps while contextual reasoning benefits from semantically richer low-resolution layers. Beyond conventional pyramid architectures, recent research has explored cross-scale re-indexing strategies to ensure that generated or retrieved features remain semantically faithful across resolution boundaries~\cite{dong2026generated}. In parallel, the computer vision community has invested considerable effort in understanding how visual representations evolve under different training regimes and architectural choices, including representation-level analyses of model behavior~\cite{yu2026vision}. The integration of multi-modal signals has also emerged as a powerful paradigm for enhancing feature discriminability: hierarchical networks guided by complementary modality-specific cues~\cite{hong2026uv} and multi-modal interest mining frameworks~\cite{zeng2026deepinterestgr} both demonstrate that fusing diverse information streams yields more robust representations. These principles extend to practical detection and tracking systems, where improved feature extraction directly translates to better behavior recognition and tracking accuracy in challenging real-world scenarios~\cite{tu2022behavior}. For SPAR, the lesson is clear: enriching the feature bank through complementary sampling paths (as in DPSConv) and fusing semantic with geometric evidence (as in MOHFE) provides a stronger foundation for subsequent query retrieval.
	
	\subsection{Candidate Retrieval and Embedding Space Optimization}
	The problem of selecting a compact set of informative candidates from a large pool shares fundamental similarities with candidate retrieval in recommender systems, where the goal is to identify a small subset of items from a massive catalog. Research in this domain has yielded valuable insights into embedding space optimization: orthogonal decomposition of embedding spaces has been shown to improve robustness by disentangling latent factors~\cite{guo2026embedding}, while factorization models provide a principled framework for understanding how latent representations support retrieval~\cite{wang2021footprint}. Recent advances in candidate-conditioned modeling~\cite{yu2026cast}, agentic context-aware reasoning~\cite{wang2026agent4poi}, and multi-task alignment with deep interest mining~\cite{zeng2026trialigngr} further demonstrate that the quality of candidate selection can be dramatically improved by incorporating richer contextual signals. Complementary strategies include prompt-based policies for cold-start scenarios~\cite{wang2025we} and graph-empowered reinforcement learning for optimization~\cite{liu2025grl}. The connection to detection is direct: just as recommender systems must filter irrelevant items from a vast candidate pool, query-based detectors must select discriminative spatial tokens from a dense feature map. HIDQ addresses this by ranking candidates according to heatmap saliency and retaining only the top-$N_q$ tokens, a strategy conceptually aligned with top-$k$ retrieval in recommendation systems. Beyond candidate selection, the broader literature on structured reasoning~\cite{mo2026dependency}, neural-symbolic inference~\cite{fu2026neurosymactive}, and anomaly-aware detection~\cite{hong2026anomaly} reinforces a key insight: incorporating structural priors and explicitly modeling outliers or noise can significantly enhance the reliability of decision-making systems. SPAR embodies this principle by treating background clutter as a form of noise to be explicitly suppressed through selective positional allocation.

	\begin{figure*}[t]
		\centering
		\includegraphics[width=\textwidth]{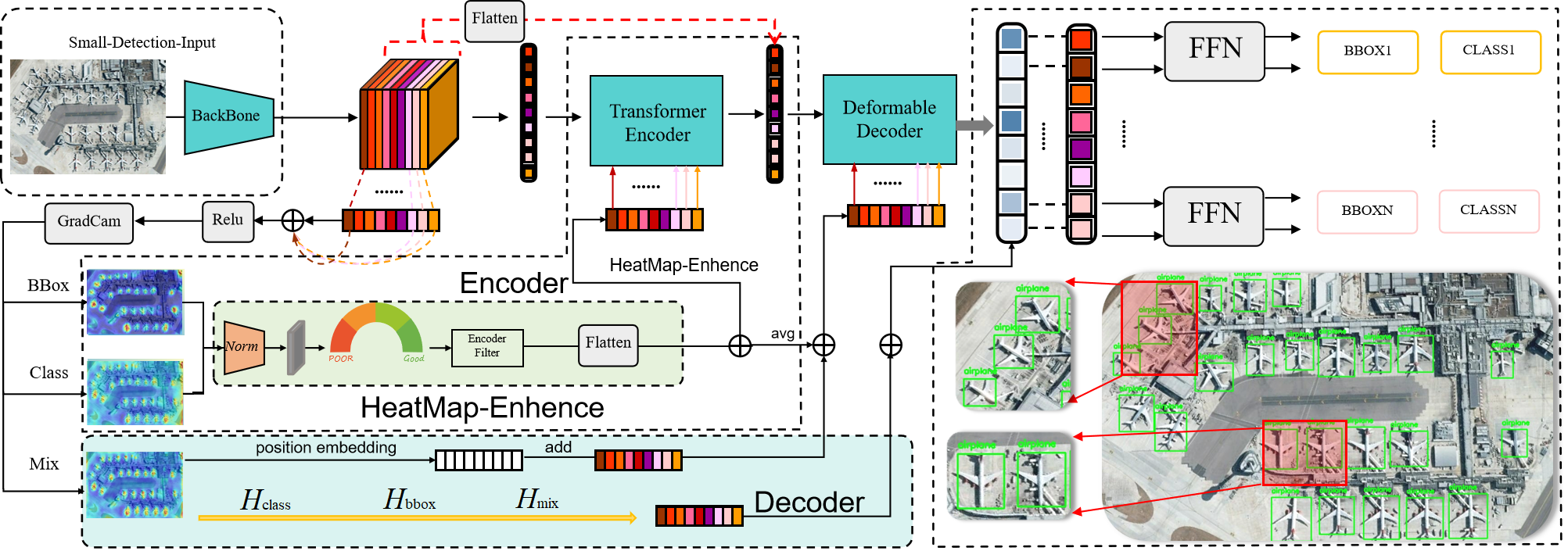}
		\caption{Overview of SPAR.
			Backbone: DPSConv enhances sparse features before transformer encoding.
			Encoder: MOHFE fuses heatmap-conditioned information with semantic and geometric cues to form cleaner multi-scale embeddings.
			Decoder: HIDQ filters background-dominant candidates with a foreground gate before query decoding.
			Bottom-right: SPE selects informative positional cues and suppresses background-dominant responses.}
		\label{fig:ppl-crop}
	\end{figure*}

	Despite these advances, transformer detectors still struggle in small-object settings because useful cues are sparse and easy to dilute in background-dominant regions. This is precisely the gap that motivates our selective positional allocation strategy.

	\section{Selective Positional Embedding}
	
	This section elaborates the SPAR component responsible for distributing positional cues in a selective, rather than uniform, manner. We begin by formalizing SPE, which transforms heatmap-derived saliency estimates into a differentiable positional gate. We subsequently detail how MOHFE and HIDQ leverage this shared gate to achieve cleaner encoding and more focused query selection. Finally, we present DPSConv as a complementary feature operator tailored to the sparse characteristics of small objects.
	
	\subsection{Selective Positional Embedding Method}
	\label{subsec:spe}
	
	Within the SPAR framework, \textbf{SPE} serves as the central allocation mechanism underpinning noise-robust query retrieval.
	As depicted in Figure~\ref{fig:ppl-crop}, SPE acquires the allocation policy by reinforcing foreground-consistent evidence while progressively dampening background-originated responses through a well-controlled three-stage procedure:
	\vspace{2mm}\noindent\emph{Step 1.} Let $\mathbf{A} \in \mathbb{R}^{K \times H \times W}$ denote an intermediate feature tensor drawn from the network, where $A_k(i, j)$ represents the activation magnitude of the $k$-th channel at spatial position $(i, j)$.
	For a target category $c$, we commence by defining a first-order saliency measure
	\begin{equation}
		s_{ijk}^{c} = \text{ReLU}\left(\frac{\partial y^c}{\partial A_k(i,j)}\right),
	\end{equation}
	which is subsequently modulated by a higher-order curvature descriptor
	\begin{equation}
		\chi_{ijk}^{c} = \left|\frac{\partial^2 y^c}{\partial A_k(i,j)^2}\right| + \eta \left|\frac{\partial^3 y^c}{\partial A_k(i,j)^3}\right|,
	\end{equation}
	with $\eta$ governing the relative contribution of second- versus third-order responses. Channel-wise significance is then obtained through a normalized aggregation scheme:
	\begin{equation}
		\beta_{k}^{c} = \frac{\sum_{i=1}^{H}\sum_{j=1}^{W} s_{ijk}^{c}\chi_{ijk}^{c}}{\sum_{i=1}^{H}\sum_{j=1}^{W} s_{ijk}^{c} + \epsilon},
	\end{equation}
	where $\epsilon$ safeguards numerical stability.
	In multi-class settings, $c$ corresponds to the ground-truth label associated with each positively matched query. To preserve computational tractability during training, Eqs.~(1)--(3) are evaluated exclusively on matched positive samples and subsequently averaged over matched instances within each mini-batch, as opposed to enumerating every category and every spatial location.

	\vspace{2mm}\noindent\emph{Step 2.} Armed with $\beta_k^c$, we construct a category-conditioned heatmap according to
	\begin{equation}
		H_{\text{class}}(i,j) = \sigma\left(\frac{1}{\sqrt{K}}\sum_{k=1}^{K}\beta_k^c A_k(i,j)\right),
	\end{equation}
	in which $\sigma(\cdot)$ denotes the logistic sigmoid nonlinearity. Analogously, by substituting the classification score $y^c$ with the regression loss $L_{\text{reg}}$, we derive a geometry-aware weighting coefficient $\beta_k^{\text{reg}}$ and assemble
	\begin{equation}
		H_{\text{bbox}}(i,j) = \sigma\left(\frac{1}{\sqrt{K}}\sum_{k=1}^{K}\beta_k^{\text{reg}} A_k(i,j)\right).
	\end{equation}
	The two heatmaps are subsequently merged into a unified saliency map via
	\begin{equation}
		H_{\text{mixed}} = \lambda H_{\text{class}} + (1-\lambda)H_{\text{bbox}}.
		\label{eqa:loss}
	\end{equation}

	\begin{figure*}[t]
		\centering
		\includegraphics[width=2\columnwidth]{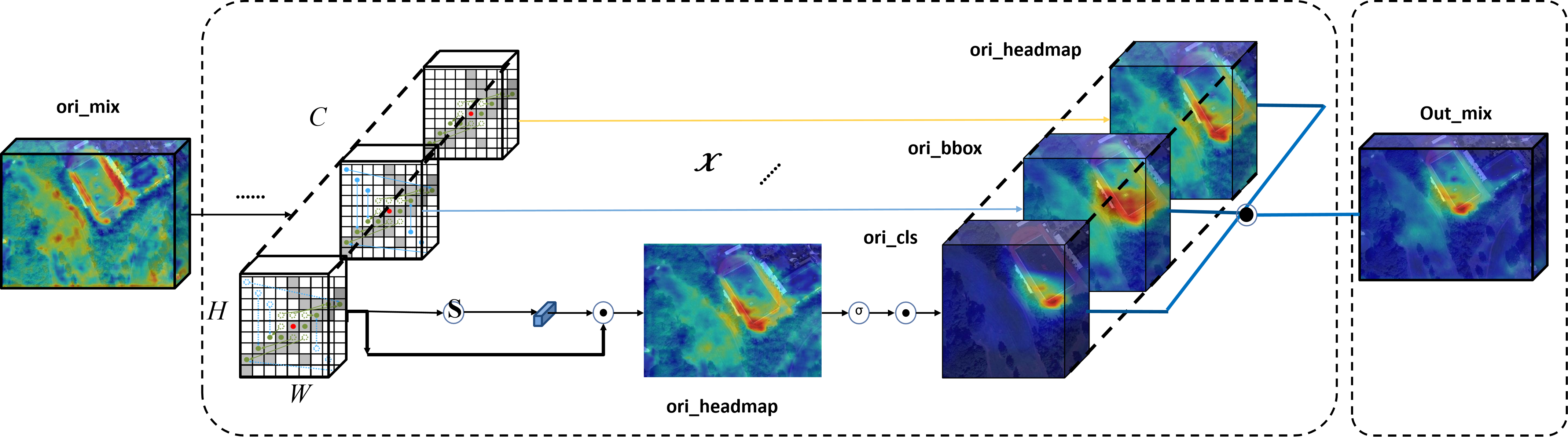}
		\caption{Detailed mechanism of the Selective Positional Embedding (SPE) module. Given an input mixed feature map (ori\_mix), SPE performs channel-wise saliency analysis augmented with higher-order curvature terms to derive class-aware (ori\_cls), box-aware (ori\_bbox), and unified heatmaps (ori\_headmap). These heatmaps are subsequently fused and thresholded to construct a spatial gate that suppresses background-dominated responses, yielding the purified output feature map (Out\_mix).}
		\label{fig:spe-mechanism}
	\end{figure*}

	\vspace{2mm}\noindent\emph{Step 3.} The unified heatmap is subsequently standardized and thresholded to yield a binary spatial gate:
	\begin{equation}
		\widehat{H}(i,j) = \frac{H_{\text{mixed}}(i,j) - \mu_H}{\sigma_H + \epsilon}, \qquad
		M_{\mathrm{hard}}(i,j) =
		\begin{cases}
			1, & \widehat{H}(i,j) > \tau, \\
			0, & \text{otherwise},
		\end{cases}
		\label{eqa:mask}
	\end{equation}
	Throughout training, we employ a straight-through estimator that retains the hard-gated forward behavior while permitting gradient flow:
	\begin{equation}
		\widetilde{M}(i,j)=\sigma\!\left(\frac{\widehat{H}(i,j)-\tau}{\gamma}\right),\quad
		M(i,j)=M_{\mathrm{hard}}(i,j)-\mathrm{sg}\!\left(\widetilde{M}(i,j)\right)+\widetilde{M}(i,j),
	\end{equation}
	where $M_{\mathrm{hard}}(i,j)=\mathbf{1}[\widehat{H}(i,j)>\tau]$, $\gamma$ denotes a temperature parameter, and $\mathrm{sg}(\cdot)$ indicates the stop-gradient operation.
	For generality, we formulate the gated positional encoding as
	\begin{equation}
		\mathrm{PE}(i,j,d) = \left[M(i,j) + \rho\bigl(1-M(i,j)\bigr)\right]\odot
		\left[\sin\left(\frac{i}{\tau_d}\right) + \cos\left(\frac{j}{\tau_d}\right)\right],
	\end{equation}
	where $\tau_d = 10000^{2d/D}$, $\rho \in [0,1)$ modulates the degree of residual leakage, and setting $\rho=0$ recovers the strict hard-suppression regime. All experiments reported herein adopt $\rho=0$.

	\vspace{2mm}
	It is worth emphasizing that heatmap-derived supervision is consumed solely during the training phase and is computed entirely from existing detection objectives, thereby obviating the need for auxiliary heatmap annotations. During inference, SPE operates as a conventional forward pass without any gradient computation, so the computational overhead associated with higher-order derivatives does not impact deployment latency. Figure~\ref{fig:spe-mechanism} encapsulates the qualitative effect: regions exhibiting strong responses retain their positional cues, whereas background-dominated zones are actively attenuated.
	The second- and third-order derivative terms are obtained via automatic differentiation restricted to selected positive queries; mixed-precision training and gradient clipping are additionally employed to enhance numerical robustness.
	
	We opt for hard gating specifically to maintain a crisp foreground/background demarcation in aerial imagery, where tiny targets possess limited spatial support and soft masking frequently leaves residual clutter unattenuated. Figure~\ref{fig:spe-mechanism} illustrates how the fused heatmap is converted into a binary gate that purifies the feature representation before it is forwarded to the encoder and decoder branches.

	\subsection{Selective Encoding and Query Retrieval}
	\label{subsec:selection}
	
	This subsection details the propagation of the SPE allocation signal through the encoder-decoder pipeline. MOHFE leverages the gated embeddings to refine multi-scale feature encoding, whereas HIDQ exploits the identical allocation signal to assemble a compact and discriminative set of decoder queries.

	\subsubsection{MOHFE: Multi-Scale ObjectBox-Heatmap Fusion Encoder}
	\label{subsubsec:MOHFE}
	
	MOHFE translates the gated positional embeddings into encoder memory that exhibits tighter correspondence with detection semantics. Its input comprises a category-conditioned embedding $E_{\text{class}}$ and a box-conditioned embedding $E_{\text{bbox}}$. The two pathways respectively accentuate category-specific evidence and geometric evidence, and MOHFE interleaves them so that both supervisory signals are preserved within the encoder state.
	
	Specifically, we first concatenate the two embeddings into $E_{\text{fuse}}=[E_{\text{class}}\|E_{\text{bbox}}]$ and subsequently project them into Query/Key/Value tensors:
	\begin{equation}
		\begin{aligned}
			Q_{\text{enc}} &= W_Q E_{\text{fuse}}, \\
			K_{\text{enc}} &= W_K E_{\text{fuse}}, \\
			V_{\text{enc}} &= W_V E_{\text{fuse}}.
		\end{aligned}
	\end{equation}
	Encoder self-attention is subsequently computed over these heatmap-modulated features. Since background-dominated positions have already been attenuated by SPE, the resulting memory representation is more compact and carries greater utility for downstream decoding.

	\subsubsection{HIDQ: Heatmap-Induced Discriminative Query Retrieval}
	\label{subsubsec:HIDQ}
	
	HIDQ transforms the unified heatmap into a concise query subset dedicated to decoding. Instead of subjecting every candidate to the decoder, it first maps the gated embedding into an initial set of query vectors:
	\begin{equation}
		Q_{\text{hidq}} = W'_Q E_{\text{mixed}}.
	\end{equation}
	These queries subsequently interact with encoder memory via deformable attention:
	\begin{equation}
		\mathit{DeformAttn}(Q_{\text{hidq}}, K_{\text{enc}}, V_{\text{enc}}).
	\end{equation}
	Candidate positions are ranked according to $H_{\text{mixed}}$, and only the top-$N_q$ tokens are retained (selected globally across all scales), where $N_q$ corresponds to the decoder query budget ($N_q=300$ under the default configuration). Decoding is performed exclusively on these selected tokens; Hungarian matching and denoising losses adhere to the standard DETR pipeline applied to the selected-query predictions.
	Because the queries are constructed from selectively allocated positional cues, they prioritize foreground-salient locations and diminish the impact of background-originated responses prior to the decoding stage.

	\begin{figure}[t]
		\centering
		\includegraphics[width=\columnwidth]{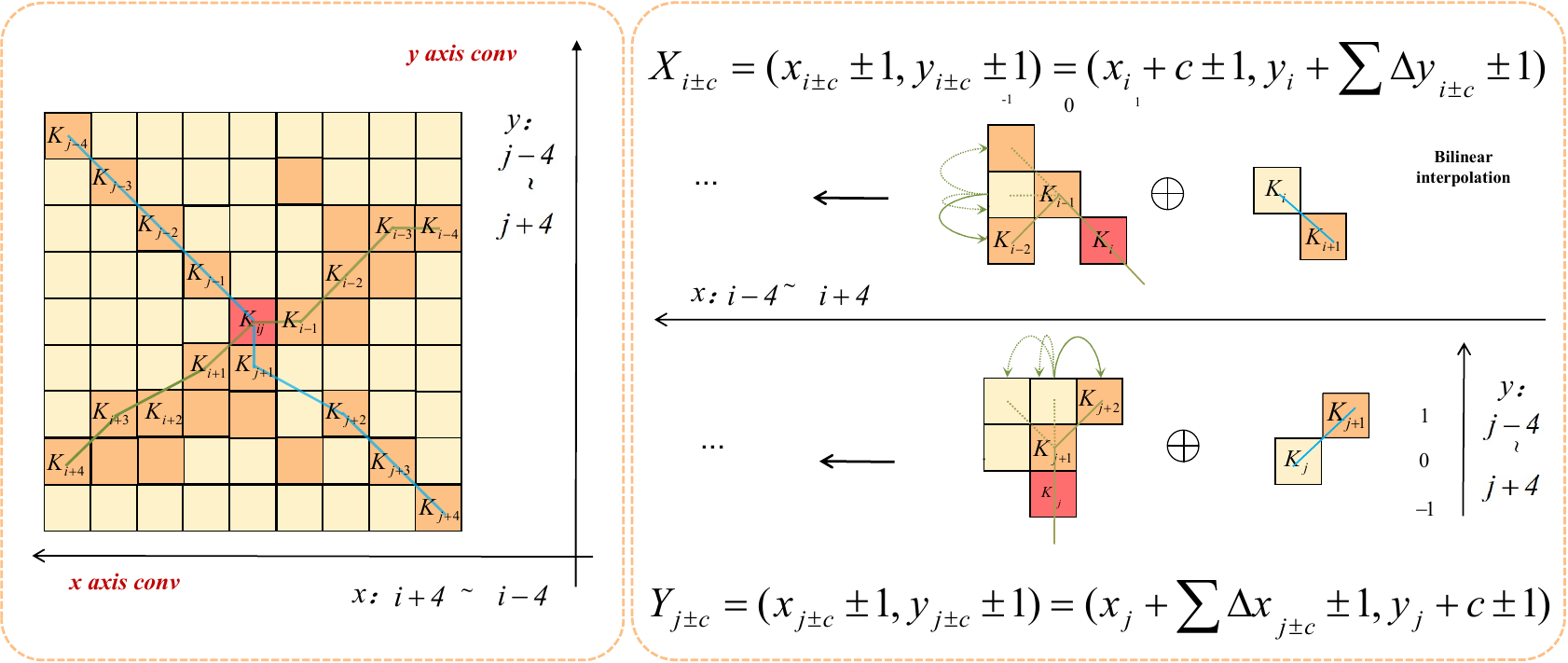}
		\caption{Dual-path axis-wise sampling in DPSConv. Left: x-/y-axis convolution samples on a $9\times 9$ grid with linear (\textcolor{blue}{blue}) and snake (\textcolor{green}{green}) paths centered at the red cell. Right: axis-wise offset accumulation and bilinear interpolation used to obtain deformed sampling locations for the two paths.}
		\label{fig:axis-crop}
	\end{figure}

	\subsection{Dual-Path Sampling Convolution}
	\label{subsec:dpsconv}
	
	Small-object detection is frequently plagued by extreme feature sparsity: discriminative evidence often manifests as slender, fragmented activations, and a solitary fixed receptive field is prone to overlooking these faint yet structured signals. To reinforce the retrieval-oriented representation prior to decoding, we introduce \textbf{DPSConv} (Dual-Path Sampling Convolution), a lightweight operator that augments local features through geometry-aware sampling while maintaining sampling stability. Given an input feature map, DPSConv produces an enhanced feature map of identical resolution, furnishing a more robust foundation for subsequent query retrieval.
	
	As illustrated in Figure~\ref{fig:axis-crop}, DPSConv employs a two-branch architecture:
	(i) a snake branch that traces irregular or curvilinear structures by means of learnable offsets, and
	(ii) a linear branch that enforces straight-line continuity through constrained sampling.
	This complementary design enables DPSConv to capture both ``non-rigid'' local evidence (advantageous for tiny, clutter-occluded objects) and ``rigid'' linear context (advantageous for structurally consistent objects), thereby yielding denser and more dependable features for the subsequent query retrieval stage.

	DPSConv estimates deformation offsets $\Delta$ via a compact convolutional predictor. To preclude unstable receptive-field drift, we regularize the offsets with a continuity constraint that ensures neighboring sampling locations evolve smoothly, thereby averting excessive deformation that could compromise feature alignment. For expository clarity, consider a $3\times 3$ kernel operating within a $9\times 9$ neighborhood. Along the horizontal axis, DPSConv employs a $3\times 1$ strip to aggregate features while stabilizing the orthogonal direction; the deformed sampling coordinates are given by:
	\begin{equation}
		\mathcal{L}_{\mathrm{cont}}=
		\sum_{p}\left(
		\left\|\Delta x_{p+1}-\Delta x_{p}\right\|_{1}+
		\left\|\Delta y_{p+1}-\Delta y_{p}\right\|_{1}
		\right),\qquad
		\mathcal{L}=\mathcal{L}_{\mathrm{det}}+\lambda_{\mathrm{cont}}\mathcal{L}_{\mathrm{cont}},
	\end{equation}
	where $\lambda_{\mathrm{cont}}$ is fixed across datasets.
	\begin{equation}
		X_{i \pm c} =
		\begin{cases}
			(x_{i+c} + 1,\; y_{i+c} + 1)
			= \left(x_i + c + 1,\; y_i + \sum \Delta y_{i+c} + 1\right), \\
			(x_{i-c} - 1,\; y_{i-c} - 1)
			= \left(x_i - c - 1,\; y_i + \sum \Delta y_{i-c} - 1\right),
		\end{cases}
	\end{equation}

	where $c \in \{0,1,2,3,4\}$ indexes the offset from the center (Figure~\ref{fig:axis-crop}, right). Feature values at these typically fractional coordinates are retrieved through bilinear interpolation, which preserves differentiability and numerical stability in DPSConv.
	
	The vertical direction is defined in an analogous manner (Figure~\ref{fig:axis-crop}, right):
	\begin{equation}
		Y_{j \pm c} =
		\begin{cases}
			(x_{j + c} + 1, y_{j + c} + 1)
			= \left(x_j + \sum \Delta x_{j + c} + 1, \; y_j + c + 1\right), \\
			(x_{j - c} - 1, y_{j - c} - 1)
			= \left(x_j + \sum \Delta x_{j - c} - 1, \; y_j + c - 1\right).
		\end{cases}
	\end{equation}

	\begin{table*}[htbp]
		\centering
		\caption{Comparison performance on PASCAL VOC and NWPU VHR-10. Compared with RT-DETR, our method achieves consistently higher accuracy on both datasets while markedly reducing computational and model complexity. The best and second-best results are highlighted in \textcolor{red}{red} and \textcolor{blue}{blue}. Entries marked with ``*'' are directly quoted from the original reports due to unavailable public training configurations; the remaining entries are reproduced in-house following each method's official recipe.}
		\label{tab:comparison}
		\renewcommand{\arraystretch}{1.2} 
		\setlength{\tabcolsep}{3pt} 
		\begin{tabular}{l l c c c c c c c c c}
			\hline \hline
			\multirow{3}{*}{ } & \multirow{3}{*}{Method} & \multirow{3}{*}{Date} & \multirow{3}{*}{Backbone} & \multicolumn{4}{c}{Datasets} & \multirow{3}{*}{Epoch} & \multirow{3}{*}{GFLOPs} & \multirow{3}{*}{\begin{tabular}{c} Params \\ (M) \end{tabular}} \\
			\cline{5-8}
			& & & & \multicolumn{2}{c}{PASCAL VOC} & \multicolumn{2}{c}{NWPU} & & & \\
			\cline{5-8}
			& & & & mAP@0.5 & mAP & mAP@0.5 & mAP & & & \\
			\hline
			\multirow{14}{*}{CNN} & SSD & 2016ECCV & RestNet18 & 51.01 & 22.98 & 40.34 & 16.01 & 75 & 353 & 26.29 \\
			& FCOS & 2019ICCV & ResNet18 & 67.41 & 43.64 & 84.35 & 55.68 & 75 & 32 & 31.8 \\
			& RetinaNet & 2018CVPR & ResNet18 & 58.21 & 44.50 & 87.78 & 57.89 &75 & 39 & 36.1 \\
			& Faster R-CNN & 2017ICCV & VGG16 & 65.97 & 38.63 & 89.15 & 61.23 &75 & 63 & 41.1 \\
			& CenterNet & 2019ICCV & ResNet34 & 57.64 & 30.90 & 81.28 & 48.76 & 75 & 130 & 41.7 \\
			& MobilenetV3 & 2019ICCV & MBNV3 & 52.40 & 30.12 & 62.97 & 40.56 & 75 & 14 & 8.86 \\
			& YOLOv3 & 2018CVPR & ResNet18 & 64.08 & 43.76 & 84.50 & 52.34 & 75 & 112 & 43.1 \\
			& YOLOv4 & 2020CVPR & Darknet-53 & 62.70 & 49.90 & 87.80 & 60.89 & 75 & 101 & 44.9 \\
			& YOLOv5 & 2020ArXiv & CSPDarknet53 & 67.10 & 49.73 & 89.11 & 65.43 & 75 & 109 & 46 \\
			\multirow{10}{*}{Transformer} & YOLOv8 & 2023Arxiv & CSPDarknet18 & 53.01 & 36.66 & 90.43 & 60.80 & 75 & 32 & 16.1 \\
			& YOLOv11 & 2025ArXiv & CSPRep53 & \textcolor{blue}{71.70} & 47.3 & \textcolor{blue}{92.68} &\textcolor{blue}{ 68.89} & 75 & 105 & 38.8 \\
			\hline
			
			& DETR & 2020ECCV & ResNet18 & 62.40 & 42.00 & 81.28 & 35.9 & 125 & 101 & 36.74 \\
			& Deformable DETR & 2021ICLR & ResNet50 & 60.56 & \textcolor{blue}{51.92} & 79.30 & 40.50 &125 & 179 & 39.83 \\
			& Conditional-DETR & 2021ICCV & ResNet50 & 68.35 & 48.72 & 91.45 & 64.38 & 100 & 98 & 240.2 \\
			& DINO & 2023ICLR & ResNet50 & 67.92 & 45.16 & 90.87 & 62.95 & 12 & 112 & 247.5 \\
			& DINOv2 & 2024ArXiv & ViT-B/14 & 69.10 & 49.85 & 92.15 & 65.70 & 24 & 189 & 301.2 \\
			& PR-Deformable DETR* & 2025GRSL & ResNet50 & / & / & 88.30 & 43.20 & 125 & 151 & 46.07 \\
			& SMWG-DETR & 2026TGRS & ResNet50 & 69.8 & 51.4 & 92.63 & 66.9 & 125 & 168 & 48.5 \\
			& RT-DETR & 2024CVPR & HGNetv1 & 69.41 & 50.59 & 92.60 & 60.26 & 100 & 136 & 163 \\
			\rowcolor{LightGreen} 
			& Our work & 2026 & SPE-Bone & \textcolor{red}{72.1} & \textcolor{red}{53.01} & \textcolor{red}{95.2} & \textcolor{red}{69.2} & 100 & 57 & 66.3 \\
			\hline \hline
		\end{tabular}
	\end{table*}

	The snake branch predominates when local evidence is fragmented or curvilinear (e.g., tiny objects embedded in clutter), since adaptive offsets can ``trace'' the activation pattern and recuperate missing contextual information. The linear branch predominates when the scene exhibits stable straight structures, since constrained sampling upholds geometric consistency and circumvents over-deformation. Figure~\ref{fig:axis-crop} depicts the two complementary sampling routes that constitute the core of DPSConv. In practice, DPSConv is embedded within the backbone to enhance feature maps prior to their delivery to the subsequent SPE-driven encoding and query retrieval modules. In summary, DPSConv offers a straightforward yet effective mechanism for densifying sparse information and enhancing geometric coherence, rendering the features better suited to selective query retrieval in small-object scenarios.

	\section{Experimental Results and Analysis}

	\subsection{Datasets and Experimental Protocol}
	Five publicly available benchmarks are employed to encompass both generic object detection and aerial small-object detection: NWPU VHR-10~\cite{cheng2016learning}, PASCAL VOC~\cite{everingham2010pascal}, DOTA, DIOR, and VisDrone. A supplementary AITOD ablation configuration is incorporated to examine model behavior under a denser aerial scene distribution. The disparities among these datasets are particularly pertinent to query-based detection, as they vary across object scale, instance density, background clutter, viewpoint diversity, and the overlap patterns exhibited by adjacent targets. Collectively, they constitute a considerably more challenging evaluation suite than any single benchmark, since a method that performs well under one scene regime may deteriorate once the background statistics or object geometry shift. Among them, NWPU VHR-10 poses the greatest challenge for positional allocation because targets are extremely small and foreground pixels account for only a minute fraction of each image. PASCAL VOC, being comparatively uncluttered, serves as an informative counterpoint for assessing model behavior in cleaner scenes. DOTA, DIOR, VisDrone, and AITOD further stress the model under large-format aerial layouts, where dense instance arrangements and pronounced scale variation render query quality and localization precision equally critical.
	
	To ensure a controlled comparison, the official training/testing splits are strictly adhered to and benchmark evaluation conventions are maintained throughout. This is crucial because the goal is to isolate the impact of selective positional allocation itself, rather than concealing improvements within dataset-specific hyper-parameter tuning. Consequently, an identical training recipe is reused wherever feasible, and only a handful of task-level constants are determined on validation data. Specifically, $\lambda=0.5$ in Eq.~\eqref{eqa:loss} balances the semantic and geometric heatmaps, while $\tau=0.5$ in Eq.~\eqref{eqa:mask} governs the sparsity of the binary gate. Following validation, both values are held constant across all experiments so that the reported metrics reflect genuine model behavior rather than per-dataset calibration.
	
	For PASCAL VOC and NWPU VHR-10, we report AP$_{50}$ and AP$_{50:95}$ (the COCO-style average computed over IoU thresholds from 0.50 to 0.95). To maintain naming consistency with prior small-object detection tables, in Table~\ref{tab:ablation} the column labeled ``mAP'' corresponds to AP$_{50}$ and ``mAP@95'' corresponds to AP$_{50:95}$. On DOTA/DIOR/VisDrone, AP$_{50}$/AP$_{75}$ is reported; DOTA is evaluated under the horizontal-box (HBB) protocol by converting oriented annotations into axis-aligned boxes. Computational efficiency is also monitored via GFLOPs and parameter count (M), since SPAR aims to enhance not only detection accuracy but also the cost of query refinement. The training strategy is kept uniform for our ablation studies and controlled comparisons; for external baselines with fixed official backbones (e.g., RT-DETR with HGNetv1), we adopt their published settings and report results without additional retuning.
	
	Unless otherwise noted, SPAR-specific hyper-parameters are fixed across all datasets: $\lambda=0.5$, $\tau=0.5$, $\gamma=0.5$, $\rho=0.0$, $N_q=300$, and $\lambda_{\mathrm{cont}}=0.05$. The remaining optimizer, data augmentation, and image resolution settings follow the official training recipes of the corresponding baseline families to ensure a controlled comparison.

	\begin{table*}[t]
		\caption{Ablation study of our design components on NWPU VHR-10, PASCAL VOC, and AITOD. Each component contributes incremental improvements, and the full configuration consistently achieves the best performance across the three datasets.}
		\label{tab:ablation}
		\adjustbox{max width=\textwidth}{
			\begin{tabular}{c c c c c|c c|c c}
				\toprule
				\toprule
				\multirow{2}{*}{Dataset} 
				& \multirow{2}{*}{DPSConv} 
				& \multirow{2}{*}{Linear-Snake} 
				& \multicolumn{2}{c|}{Selective Positional Embedding} 
				& \multirow{2}{*}{Precision(\%)} 
				& \multirow{2}{*}{Recall(\%)} 
				& \multirow{2}{*}{mAP(\%)} 
				& \multirow{2}{*}{mAP@95(\%)} \\
				\cmidrule(lr){4-5}
				& & & MOHFE & HIDQ & & & & \\
				\midrule
				
				\multirow{5}{*}{NWPU VHR-10}
				& - & - & - & - 
				& 88.9 & 89.9 & 92.6 & 60.2 \\
				& \ding{51} & - & - & - 
				& 89.0$_{\uparrow 0.10}$ & 90.10$_{\uparrow 0.20}$ & 92.71$_{\uparrow 0.11}$ & 62.32$_{\uparrow 2.12}$ \\
				& \ding{51} & \ding{51} & - & - 
				& 90.74$_{\uparrow 1.84}$ & 90.62$_{\uparrow 0.72}$ & 93.91$_{\uparrow 1.31}$ & 63.69$_{\uparrow 3.49}$ \\
				& \ding{51} & \ding{51} & \ding{51} & - 
				& 90.75$_{\uparrow 1.85}$ & 90.51$_{\uparrow 0.61}$ & 94.18$_{\uparrow 1.58}$ & 63.94$_{\uparrow 3.74}$ \\
				\rowcolor{LightGreen}
				& \ding{51} & \ding{51} & \ding{51} & \ding{51} 
				& 93.17$_{\uparrow 4.97}$ & 90.40$_{\uparrow 0.50}$ & 95.2$_{\uparrow 2.60}$ & 69.2$_{\uparrow 9.00}$ \\
				\midrule
				
				\multirow{5}{*}{PASCAL VOC}
				& - & - & - & - 
				& 76.7 & 62.9 & 69.4 & 50.5 \\
				& \ding{51} & - & - & - 
				& 77.20$_{\uparrow 0.50}$ & 62.93$_{\uparrow 0.03}$ & 70.29$_{\uparrow 0.89}$ & 51.17$_{\uparrow 0.67}$ \\
				& \ding{51} & \ding{51} & - & - 
				& 78.41$_{\uparrow 1.71}$ & 63.10$_{\uparrow 0.20}$ & 70.19$_{\uparrow 0.79}$ & 51.41$_{\uparrow 0.91}$ \\
				& \ding{51} & \ding{51} & \ding{51} & - 
				& 78.52$_{\uparrow 1.82}$ & 63.55$_{\uparrow 0.65}$ & 70.47$_{\uparrow 1.07}$ & 51.33$_{\uparrow 0.83}$ \\
				\rowcolor{LightGreen}
				& \ding{51} & \ding{51} & \ding{51} & \ding{51} 
				& 78.80$_{\uparrow 2.10}$ & 63.72$_{\uparrow 0.82}$ & 72.1$_{\uparrow 2.70}$ & 53.01$_{\uparrow 2.51}$ \\
				\midrule
				
				\multirow{5}{*}{AI-TOD}
				& - & - & - & - 
				& 61.2 & 55.0 & 49.8 & 17.8 \\
				& \ding{51} & - & - & - 
				& 61.4$_{\uparrow 0.20}$ & 55.2$_{\uparrow 0.20}$ & 50.0$_{\uparrow 0.20}$ & 18.0$_{\uparrow 0.20}$ \\
				& \ding{51} & \ding{51} & - & - 
				& 62.1$_{\uparrow 0.90}$ & 55.8$_{\uparrow 0.80}$ & 50.5$_{\uparrow 0.70}$ & 18.4$_{\uparrow 0.60}$ \\
				& \ding{51} & \ding{51} & \ding{51} & - 
				& 62.4$_{\uparrow 1.20}$ & 56.1$_{\uparrow 1.10}$ & 50.8$_{\uparrow 1.00}$ & 18.6$_{\uparrow 0.80}$ \\
				\rowcolor{LightGreen}
				& \ding{51} & \ding{51} & \ding{51} & \ding{51} 
				& 63.0$_{\uparrow 1.80}$ & 56.8$_{\uparrow 1.80}$ & 51.4$_{\uparrow 1.60}$ & 19.0$_{\uparrow 1.20}$ \\
				
				\bottomrule
				\bottomrule
			\end{tabular}
		}
	\end{table*}

	\begin{figure*}[t] 
		\centering
		\includegraphics[width=0.76\textwidth]{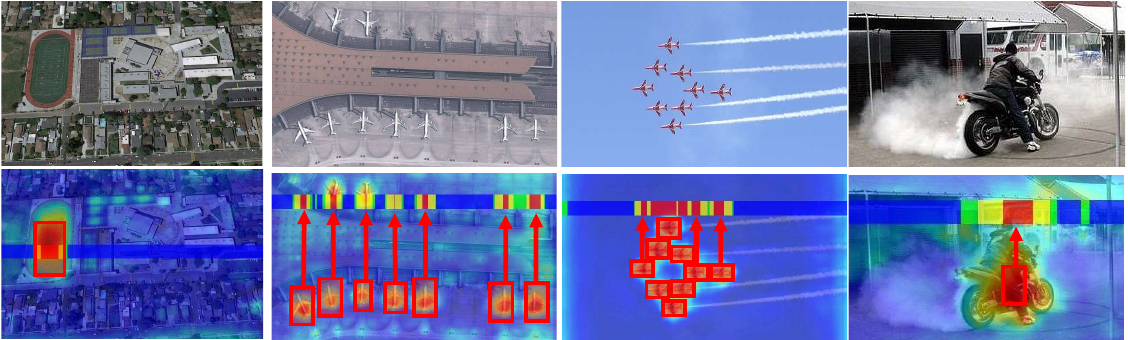}
		\caption{Visualization of SPE activation bars under cluttered versus clean scenes. The selective allocation provides more object-aware spatial information, leading to tighter and more stable localization under background distractions and scale variations.}
		\label{fig:vis2}
	\end{figure*}

	\subsection{Result Analysis}
	\subsubsection{Quantitative Results}
	The comparison ensemble encompasses representative CNN-based detectors, specifically FCOS~\cite{tian2019fcos}, RetinaNet~\cite{lin2017focal}, Faster R-CNN~\cite{ren2015faster}, CenterNet~\cite{duan2019centernet}, MobileNet-V3~\cite{howard2019searching}, and the YOLO series~\cite{redmon2018yolov3,bochkovskiy2020yolov4,benjumea2021yolo,li2022yolov6,wang2023yolov7,khanam2024yolov11}. Transformer-based baselines are likewise incorporated, including DETR, Deformable DETR, PR-Deformable DETR*, and RT-DETR, as they constitute the most pertinent reference group for a decoder-centric retrieval design. To preserve fairness, every baseline is evaluated under its canonical training recipe or a widely adopted schedule, and no dataset-specific manual tuning is introduced beyond the protocol delineated above.
	
	Table~\ref{tab:comparison} unveils a consistent trend across both benchmark categories. CNN-based methods exhibit competitive convergence speed, yet their accuracy is bounded by the computational overhead of dense proposal processing and the discrepancy between candidate generation and final selection. Transformer detectors eliminate NMS and adopt end-to-end set prediction, yielding a structurally cleaner pipeline; however, their efficacy is heavily contingent upon the quality of the initial queries. This limitation is particularly pronounced on NWPU VHR-10, where objects are minuscule, background regions predominate, and query initialization must contend with exceedingly feeble spatial evidence. Under such conditions, DETR lags substantially behind the stronger methods, while RT-DETR, despite being considerably more capable, still leaves scope for a more selective allocation strategy.
	
	The head-to-head comparison with RT-DETR is particularly revealing. On NWPU VHR-10, SPAR attains 95.2\% mAP@0.5 and 69.2\% mAP while concurrently lowering computational cost from 136 to 57 GFLOPs and parameter count from 163M to 66.3M. The observation that the improvement is more pronounced on mAP than on mAP@0.5 is not coincidental: it signifies that the model is enhancing the geometric precision of predictions rather than merely expanding detection coverage. Put differently, the gains are concentrated at stricter localization thresholds, which is precisely the regime where selective positional allocation ought to exert the greatest influence. 
	
	On PASCAL VOC, an analogous trend emerges, albeit with smaller absolute gains owing to the dataset's reduced clutter and less severe small-object regime. SPAR achieves 72.1\% mAP@0.5 and 53.01\% mAP on this benchmark, surpassing RT-DETR by +2.69 and +2.42 points, respectively. Despite the narrower margin, the improvement remains consistent across both accuracy and efficiency metrics, which is significant because it demonstrates that SPAR is not overfit to a particular aerial texture distribution. Rather, the method appears to embody a transferable principle: when query quality is enhanced upstream, the decoder can allocate less capacity to noise correction and more capacity to refining meaningful predictions. For this reason, the reported improvements should be interpreted not merely as higher scores, but as evidence that selective positional allocation constitutes a more efficient utilization of model capacity.
	
	AITOD furnishes a denser aerial setting for the ablation investigation. On this benchmark, the complete configuration reaches 51.4\% mAP and 19.0\% mAP@95, up from 49.8\% and 17.8\% in the baseline. Although the gains are more modest than those observed on NWPU VHR-10, they remain consistent across all constituent modules, which suggests that the proposed allocation-and-retrieval pipeline continues to enhance fine-grained localization even as object density and inter-instance overlap increase.

	\subsubsection{Qualitative Visualization}
	
	Figure~\ref{fig:vis2} depicts the activation bars generated by SPE on PASCAL VOC and NWPU VHR-10. The activation patterns are compact rather than diffuse: intense responses cluster around the object body, while surrounding clutter is rapidly attenuated. This behavior is noteworthy because it implies that SPE is not merely acquiring a generic saliency map. Instead, it is executing a more specialized allocation operation, retaining positional cues where they facilitate decoding and suppressing them where they are apt to introduce ambiguity.
	
	The downstream implications become more apparent when detector behavior is taken into account. When the activation profile is centered and sharp, the query source inherits a cleaner spatial prior, which renders the decoder less susceptible to nearby distractors and less prone to drift under resizing or padding operations. As scene clutter intensifies, the distinction between active and suppressed regions becomes even more consequential, because the gate precisely eliminates the regions that would otherwise inject unstable positional information into the query set. Viewed from this perspective, the visualization is not merely a qualitative illustration; it constitutes direct evidence that the model is acquiring an allocation policy rather than a passive highlighting mechanism.
	
	Across both clean and congested scenes, the activation bars retain their interpretability. In relatively straightforward images, SPE tends to preserve a tighter foreground core, whereas in more challenging aerial scenes it expands the active region only when additional contextual information is required for reliable localization. This adaptive behavior is consistent with SPAR's role as a selective allocation framework: the method does not compel every location to contribute equally, but instead modulates the positional prior in accordance with the retrieval utility of each region.

	\subsection{Ablation Studies and Additional Evaluation}
	
	\subsubsection{Component-wise ablation}
	Table~\ref{tab:ablation} disassembles SPAR into its constituent stages and demonstrates that the performance gains are cumulative rather than redundant. This is significant because the three modules operate at distinct junctures within the pipeline: DPSConv modulates the quality of the local feature bank, MOHFE restructures the encoder-side memory, and HIDQ filters the decoder-side query source. If the initial stage is deficient, subsequent stages have scant structure to exploit; if the encoder memory remains noisy, the decoder continues to be exposed to background interference; and if the query set is not purified prior to decoding, the final predictor must expend capacity rectifying avoidable errors. The ablation thus corroborates the notion that SPAR functions as a coordinated allocation-and-retrieval system rather than as an isolated heuristic.
	
	\noindent\textbf{NWPU VHR-10.}
	The baseline attains 92.6\% mAP@0.5 and 60.2\% mAP. Incorporating the initial dual-path sampler yields a modest yet dependable improvement, consistent with the intuition that diminutive aerial targets profit from more adaptive local aggregation. When the complete DPSConv design is activated, the gain becomes more evident at stricter localization thresholds, indicating that the module is not simply augmenting feature density but also preserving geometric coherence in the vicinity of small objects. MOHFE subsequently furnishes a cleaner encoder prior by infusing the gated positional signals into multi-scale memory, which tends to enhance the detector's capacity to anchor foreground regions prior to decoding. HIDQ delivers the most substantial final increment because it eliminates background-dominated candidates at the stage where query noise incurs the highest cost. 
	
	\noindent\textbf{PASCAL VOC.}
	On PASCAL VOC, an identical progression manifests, but with smaller absolute gains due to the easier scene distribution and lower clutter intensity. The final configuration attains 72.1\% mAP@0.5 and 53.01\% mAP, which is consistent with the improvements documented in Table~\ref{tab:comparison}. Even so, the complete configuration still enhances both accuracy metrics consistently, which is significant for two reasons. First, it demonstrates that SPAR is not specialized to aerial imagery or to exceedingly dense scenes. Second, it suggests that selective positional allocation constitutes a more general mechanism for query-based detection whenever the model must discriminate foreground evidence from surrounding background structure.
	
	\noindent\textbf{AITOD.}
	AITOD introduces a denser aerial setting in which instances overlap more frequently and background suppression assumes greater importance. The complete configuration reaches 51.4\% mAP and 19.0\% mAP@95, surpassing the baseline by 1.6 and 1.2 points, respectively. Although the absolute gains are more modest than those on NWPU VHR-10, they remain stable across all stages, which indicates that the proposed modules continue to provide benefits when the scene is congested and the localization target is more stringent.

	\subsubsection{Ablation on DETR decoder depth}

	\begin{table}[t]
		\setlength{\aboverulesep}{0.5ex}
		\setlength{\belowrulesep}{0.5ex}
		\renewcommand{\arraystretch}{1.2}
		
		\caption{Decoder multi-head ablation under different training regimes. Det$^k$ denotes using $k$ decoder layers for detection, while Det$^{3\text{--}7}$ aggregates outputs from layers 3 to 7, where performance saturates. Results show that increasing decoder depth consistently improves AP under limited training budgets, but exhibits diminishing returns beyond two layers under sufficient training.}
		\label{tab:params}
		\centering
		
		\resizebox{\linewidth}{!}{%
			\begin{tabular}{c|c c c c|c c c c}
				\toprule
				\toprule
				\multirow{2}{*}{Epoch} & \multicolumn{4}{c}{AP(\%)} & \multicolumn{4}{c}{GFLOPs} \\
				\cline{2-5} \cline{6-9}
				& Det$^0$ & Det$^1$ & Det$^2$ & Det$^{3\text{--}7}$ & 0 & 1 & 2 & 3\text{--}7 \\
				\midrule
				100 & 66.9 & 70.3 & \textcolor{red}{70.6} & 70.6
				& \multirow{4}{*}{53.7} & \multirow{4}{*}{55.4} & \multirow{4}{*}{56.2} & \multirow{4}{*}{57.0} \\
				50  & 62.6 & 65.7 & 63.2 & \textcolor{red}{66.9} &  &  &  &  \\
				25  & 40.6 & 41.4 & 43.3 & \textcolor{red}{48.0} &  &  &  &  \\
				Early-stop & 39.2 & 43.0 & 43.9 & \textcolor{red}{44.0} &  &  &  &  \\
				\bottomrule
				\bottomrule
			\end{tabular}%
		}
	\end{table}

	Table~\ref{tab:params} isolates the effect of decoder depth while holding SPE-driven query construction constant. The observed trend is illuminating because it reveals how the decoder's role evolves once query quality is enhanced upstream. Under the complete training schedule, Det$^{2}$ already attains 70.6\% AP, which signifies that an extended refinement chain is no longer indispensable for recovering from noisy initial queries. Conversely, when optimization is restricted to fewer epochs, deeper decoders remain beneficial because they can partially compensate for representations that have not yet fully converged. In essence, decoder depth retains utility as a fallback mechanism, but its marginal contribution diminishes once the preceding allocation stage becomes effective.
	
	This behavior aligns with the underlying design philosophy of SPAR. Instead of relying on depth to iteratively rectify suboptimal inputs, the method shifts greater responsibility to the query construction stage so that the decoder commences from a cleaner initial state. Once this is achieved, removing additional layers does not precipitate model collapse, because the refinement task itself becomes more tractable. The modest variation in GFLOPs across decoder depths also underscores a notable asymmetry in transformer detectors: the decoder is not invariably the dominant source of floating-point operations, but it is frequently the primary locus where parameters are allocated to correction and refinement. SPAR consequently enhances not only prediction accuracy, but also the distribution of model capacity throughout the pipeline.

	\subsubsection{Additional Evaluation on Benchmarks}
	\label{subsec:large-scale}

	\begin{table}[t]
		\centering
		\caption{Generalization to large-scale aerial benchmarks. We report AP at IoU thresholds 0.50 and 0.75 (AP$_{50}$ / AP$_{75}$). ``Ours'' consistently improves over the DETR across all datasets.}
		\label{tab:aerial}
		\renewcommand{\arraystretch}{1.3}
		\resizebox{0.9\columnwidth}{!}{
			\begin{tabular}{l c c c c c c}
				\toprule
				\toprule
				\multirow{2}{*}{Method} 
				& \multicolumn{2}{c}{DOTA} 
				& \multicolumn{2}{c}{DIOR} 
				& \multicolumn{2}{c}{VisDrone} \\
				\cmidrule(lr){2-3} \cmidrule(lr){4-5} \cmidrule(lr){6-7}
				& AP$_{50}$ & AP$_{75}$ 
				& AP$_{50}$ & AP$_{75}$ 
				& AP$_{50}$ & AP$_{75}$ \\
				\midrule
				DETR & 33.7 & 15.4 & 56.2 & 25.8 & 42.1 & 19.7 \\
				\rowcolor{LightGreen} 
				Ours & 40.2$_{\uparrow 6.5}$ & 19.4$_{\uparrow 4.0}$ 
				& 63.9$_{\uparrow 7.7}$ & 31.3$_{\uparrow 5.5}$ 
				& 49.7$_{\uparrow 7.6}$ & 24.2$_{\uparrow 4.5}$ \\
				\bottomrule
				\bottomrule
			\end{tabular}
		}
	\end{table}
	
	Generalization capability is further investigated on DOTA, DIOR, and VisDrone through comparison with a DETR baseline. The improvements are uniform across all three benchmarks and at both IoU thresholds, which is significant because AP$_{75}$ is considerably more stringent than AP$_{50}$ and consequently more sensitive to localization precision. The observation that SPAR enhances both metrics implies that the method is not merely detecting a greater number of objects, but also positioning them with tighter bounding boxes. In the updated results, the model achieves 40.2/19.4 on DOTA, 63.9/31.3 on DIOR, and 49.7/24.2 on VisDrone for AP$_{50}$/AP$_{75}$, preserving the same overall trend while delivering a small yet consistent uplift. This distinction is particularly relevant in aerial detection, where coarse localization can still produce a moderate AP$_{50}$ score while failing under stricter overlap criteria.
	
	The benchmark-specific behavior is likewise instructive. DOTA encompasses numerous small and oriented objects within large-format scenes, DIOR covers a diverse array of remote-sensing categories with heterogeneous instance sizes, and VisDrone introduces severe occlusion and congested street-level imagery. SPAR remains advantageous across all three settings, which implies that its allocation mechanism is not contingent upon a particular object geometry or background texture pattern. Rather, the model appears to derive benefits whenever the detection task necessitates separating sparse foreground evidence from a dense and noisy contextual background.

	\section{Conclusion}
	
	A simple but under-explored question motivates the conclusion of this work: how should positional cues be allocated so that the detector preserves useful structure while suppressing background noise? In cluttered scenes, indiscriminate positional injection allows background regions to contribute distracting signals, which weakens retrieval and forces the decoder to spend extra capacity correcting noisy queries. SPAR addresses this by making positional allocation selective: foreground-salient regions retain stronger positional cues, while background-dominant regions are attenuated. The result is cleaner encoder memory, more reliable decoder queries, and a detector that can be made substantially lighter without sacrificing performance.
	
	Taken together, the experiments show that this is not just an embedding tweak but a pipeline-level change in how model capacity is used. By improving the quality of the query source, SPAR reduces the need for deep refinement chains and yields a better balance between accuracy and efficiency. Future work includes exploring prediction-driven alternatives to gradient-based allocation and studying how SPE transfers to other query-based detection architectures and broader vision settings.

	\clearpage
	
	\bibliographystyle{ACM-Reference-Format}
	\bibliography{refs}

\end{document}